\documentclass[10pt,twocolumn,letterpaper]{article}

\usepackage{cvpr}              

\usepackage[accsupp]{axessibility}  

\usepackage[dvipsnames]{xcolor}

\definecolor{cvprblue}{rgb}{0.21,0.49,0.74}
\usepackage[pagebackref,breaklinks,colorlinks,citecolor=cvprblue]{hyperref}

\usepackage{algorithm}
\usepackage[noend]{algpseudocode}
\usepackage[frozencache]{minted}

\DeclareMathOperator*{\argmax}{\arg\,\max}

\def\paperID{10}
\def\confName{CVPR}
\def\confYear{2024}

\def\pkg{\emph{SGBench}}

\title{A Review and Efficient Implementation of Scene Graph Generation Metrics}

\author{Julian Lorenz \qquad Robin Schön \qquad Katja Ludwig \qquad Rainer Lienhart\\
University of Augsburg, Germany\\
{\tt\small \{julian.lorenz,robin.schoen,katja.ludwig,rainer.lienhart\}@uni-a.de}
}

\begin{document}
\maketitle


\begin{abstract}
Scene graph generation has emerged as a prominent research field in computer vision, witnessing significant advancements in the recent years.
However, despite these strides, precise and thorough definitions for the metrics used to evaluate scene graph generation models are lacking.
In this paper, we address this gap in the literature by providing a review and precise definition of commonly used metrics in scene graph generation. Our comprehensive examination clarifies the underlying principles of these metrics and can serve as a reference or introduction to scene graph metrics.

Furthermore, to facilitate the usage of these metrics, we introduce a standalone Python package called \pkg{} that efficiently implements all defined metrics, ensuring their accessibility to the research community.
Additionally, we present a scene graph benchmarking web service, that enables researchers to compare scene graph generation methods and increase visibility of new methods in a central place.

All of our code can be found at \url{https://lorjul.github.io/sgbench/}.
\end{abstract}

\section{Introduction}

Scene graph generation (SGG) aims to represent images as graphs where nodes correspond to objects in the image and edges denote relationships or interactions between the objects.
The quality of generated scene graphs are pivotal in various downstream tasks, but despite the growing interest and advancements in scene graph generation models, evaluation protocols have been lacking formal definitions for the used metrics and rely on implementation details.
In this paper, we address this issue by proposing comprehensive and formal definitions for scene graph generation metrics, providing a solid foundation for benchmarking and thus helping to advance research in this domain.

Many scene graph papers introduce the used metrics but don't explicitly define them. Even most survey papers on scene graph generation do not cover the metrics thoroughly. In contrast, we put much focus on a precise language to define commonly used scene graph metrics. We provide formal definitions and accompanying pseudo code to strictly describe Recall@k, Mean Recall@k, Pair Recall@k, and more. This paper thus serves as a reference and introduction to scene graph metrics. Finally, we provide evaluation results for existing panoptic scene graph methods with the discussed metrics.

In addition, we provide an efficient Python implementation of the discussed metrics. Our implementation is designed to run faster and use less disk space. On top, we use less boilerplate code and reduce the total number of lines of code from more than 1500 down to less than 700 compared to OpenPSG \cite{psg}. Our implementation is designed to be easy to understand and adaptable for future needs. All of our code is available here: \url{https://lorjul.github.io/sgbench/}.

To further advance the field of scene graph generation and improve visibility of new scene graph methods, we additionally introduce a public benchmarking web service. We aim to establish this service as a central place to compare new scene graph methods for various tasks and datasets.

To summarize, our contributions are:

\begin{enumerate}
    \item A thorough review of commonly used metrics in scene graph generation with precise definitions.
    \item A comparison of existing panoptic scene graph methods on the discussed metrics.
    \item An efficient Python package called \pkg{} that is lightweight, easy to use, and supports all discussed metrics.
    \item A benchmarking service where scene graph models can be evaluated and compared.
\end{enumerate}

\section{Related Work}


\subsection{Metrics}

The first scene graph metric Recall@k was introduced by \citeauthor{first_scenegraph} \cite{first_scenegraph}.
Since then many surveys have discussed the topic of scene graph generation. However, most of them do not define the used scene graph metrics.
\citeauthor{survey2023} \cite{survey2023} compare Recall@k, Mean Recall@k, and No Graph Constraint Recall@k. However, they limit the section for the metrics to a single paragraph without a thorough definition and focus more on different scene graph architectures and applications. \citeauthor{survey2022} \cite{survey2022} only loosely define Recall@k and mention Mean Recall@k just one single time. \citeauthor{survey2020} \cite{survey2020} verbally define Recall@k, Mean Recall@k, and No Graph Constraint Recall@k but don't provide any strict definitions. \citeauthor{survey2024} \cite{survey2024} are more precise than the previously mentioned surveys. They mathematically define Recall@k, Mean Recall@k, and No Graph Constraint Recall@k but do not dive into the details as much as we do in this paper.
In addition, they define Zero-Shot Recall@k as the Recall@k for relation triplets that are not in the training set. Since this metric is just Recall@k with a different test set, we do not treat it as a separate metric in this paper.

Although most scene graph surveys loosly define the used metrics, they do not dive into the details.
In this paper, we will formalize commonly used scene graph metrics and provide explicit pseudo code and thorough mathematical definitions. In addition we provide a reference implementation that was written from the ground up to be easy to use and customize.


\subsection{Scene Graph Models}

In the original scene graph generation task (SGG), a scene graph consists of relations between bounding boxes \cite{visual_genome}. An extension to this approach is panoptic scene graph generation (PSGG) \cite{psg}, where segmentation masks are used instead of bounding boxes. All the discussed metrics in this paper apply to both SGG and PSGG and our \pkg{} package supports both tasks. To keep this paper concise, we choose PSGG as an example to demonstrate the discussed metrics and our implementation. We argue that PSGG is a more challenging and meaningful task than SGG.

In the first paper for PSGG \cite{psg}, the following methods have been ported to the panoptic task: IMP \cite{imp}, Neural-Motifs \cite{motifs}, GPS-Net \cite{gpsnet}, and VCTree \cite{vctree}. These methods were originally designed for the SGG task but adapted for the PSGG task.
PSGTR \cite{psg} and PSGFormer \cite{psg} are end-to-end scene graph architectures that are built on DETR \cite{detr} and HOTR \cite{hotr}.
Pair-Net \cite{pairnet} is an extension of PSGFormer and splits the prediction step into pair detection and then predicate classification.
HiLo \cite{hilo} is another one-stage scene graph method that builds on the ideas from \cite{ietrans}. For each relation it predicts two predicate distributions. One for low frequency predicates and one for high frequency predicates. The distributions are then merged to one single distribution to tackle the predicate imbalance issue with scene graph datasets.

For all these models, we provide conversion scripts that can be used to convert the model-specific output to our generalized format, discussed in \cref{sec:format}. Future methods that build on these models can use the same conversion scripts with virtually no additional effort.

\section{Terminology}

To better understand the metric definitions, we define the terminology here that is used throughout this paper.

\paragraph{Instances}
An instance refers to a visual object in an image. It is identified by a segmentation mask (or bounding box) and a class label. We define $M_{gt}$ as the set of ground truth instances in an image and $M_{out}$ as the set of predicted instances in an image. We use the term "instances" instead of "objects" to avoid confusion with subjects/objects on relations.

\paragraph{Predicate}
A predicate is a single label that can be used to describe a relation between a subject instance and an object instance (\eg, "holding", "looking at", "parked on", \dots). We define $P$ as the set that contains all possible predicate classes. The PSG dataset \cite{psg} that we use in \cref{sec:psgg_results} contains 56 elements in $P$.

\paragraph{Relation Triplet}
We define a relation triplet (or triplet for short) as a 3-tuple of a subject instance, a predicate label, and an object instance. Note that a triplet does not contain any scores for the predicate, the subject, or the object. If it is a ground truth triplet, we define it as $(sbj, predicate, obj) \in M_{gt} \times P \times M_{gt}$. If the triplet was returned by a scene graph model, we define it as $(sbj, predicate, obj) \in M_{out} \times P \times M_{out}$. For a triplet $t$, we use $t_{sbj}$ to refer to the related subject. $t_{obj}$ and $t_{predicate}$ are used respectively.

\paragraph{Scene Graph}
A scene graph is a graph representation that encodes interactions between visual objects in an image. It consists of a set of instances that are connected by a set of relations, defined as relation triplets.

\section{Metrics}


In this section, we formalize the commonly used set of metrics for scene graph generation. In contrast to existing definitions, we define them more formally and provide pseudo code for a better understanding.

\subsection{Converting Scores to Triplets}
\label{sec:convert_triplets}

Existing metrics implementations do not receive a sequence of triplets as input but directly operate on the estimated predicate confidence scores. This approach removes the flexibility that a model can decide on its own how to sort the triplets. Therefore, we decide to strictly decouple the triplet conversion step from the definition of the metrics.

The confidence scores can be converted to an ordered sequence of subject-predicate-object triplets that are used for the discussed metrics. Let $s$ be the confidence score for a predicate $p$ on a relation with subject $sbj$ and object $obj$. Then, $(sbj, p, obj)$ form a triplet that can be sorted by its score $s$.
We provide scripts that perform this conversion step for all discussed scene graph methods.

\subsection{General Structure}

Given a sequence of triplets, all scene graph metrics follow a similar scheme, depicted in \cref{alg:overview}.

\textbf{(1)} A subset of the triplets is selected. This depends on the used metric and is discussed in the respective sections. For example, most metrics have a $k$ parameter that defines how many predicted triplets are allowed per image.
\textbf{(2)} The selected triplets contain references to the predicted instances. To calculate metric scores, these references have to be replaced by references to the ground truth instances. Therefore, predicted instances are matched to ground truth instances, as discussed in \cref{sec:inst_match}. In this process, some triplets contain references to instances that cannot be matched. These triplets are removed from the selection.
\textbf{(3)} The matched triplets can be compared to the ground truth triplets and a score for the image can be calculated. The exact calculation depends on the used metric and is discussed in the respective sections.
\textbf{(4)} All scores for the individual images are averaged to calculate the score for the whole dataset.

Steps 2 and 4 are the same for every metric. Therefore, we define each metric by their triplet selection and score calculation in \cref{sec:rk,sec:mrk,sec:prk,sec:ngrk,sec:mngrk}.

\begin{algorithm}[tb]
	\caption{General Metric Framework}
	\label{alg:overview}
	\begin{algorithmic}[1]
		\State \textbf{Input:} Output triplets $X_i$ in descending order from a scene graph model for each image $i$ in the dataset, ground truth triplets $G_i$ for each image, metric function $f$ at image-level
		\State \textbf{Output:} Metric score for the whole dataset
		\Procedure{Framework}{$f, G, X$}
			\State $S \gets \emptyset$
			\ForAll{images $i$ in the dataset}
				\State $X' \gets$ relevant subset of $X_i$
				\State $L \gets GetMapping()$ \Comment \cref{alg:get_mapping}
				\State $X' \gets ApplyMatching(L, X')$ \Comment \cref{alg:apply_match}
				\State $s \gets f(G_i, X')$ \Comment Calculate metric
				\State $S \gets S \cup \{s\}$
			\EndFor
			\State \textbf{return} $\overline{S}$ \Comment Average of $S$
		\EndProcedure
	\end{algorithmic}
\end{algorithm}

\subsection{Instance Matching}
\label{sec:inst_match}

\begin{algorithm}[tb]
	\caption{Create Mapping for Instance Matching}
	\label{alg:get_mapping}
	\begin{algorithmic}[1]
		\State \textbf{Input:} Predicted instances $M_{out}$, ground truth instances $M_{gt}$, and minimum IoU threshold $t$
		\State \textbf{Output:} Mapping $L$ that maps instances from $M_{out}$ to instances in $M_{gt}$
		\Procedure{GetMapping}{$M_{out}, M_{gt}, t$}
		\State Initialize mapping $J$ with $J[x] = null$ $\forall x \in M_{gt}$
		\ForAll{$m$ in $M_{out}$}
			\State $M'_{gt} \gets$ subset of $M_{gt}$ that contains only instances with the same class label as $m$
			\State $x \gets \argmax\limits_{g \in M'_{gt}}\  iou(g, m)$
			\If {$iou(x, m) > t$}
			\If {$J[x] = null \text{ or } iou(x, m) > iou(x, J[x])$}
				\State $J[x] \gets m$
			\EndIf
			\EndIf
		\EndFor
		\State $L \gets J^{-1}$ \Comment Use the inverse mapping of $J$
		\State \textbf{return} $L$
		\EndProcedure
	\end{algorithmic}
\end{algorithm}

\begin{algorithm}[tb]
	\caption{Apply Instance Matching}
	\label{alg:apply_match}
	\begin{algorithmic}[1]
		\State \textbf{Input:} Mapping $L{:\,} M_{out} \rightarrow M_{gt}$ (\cref{alg:get_mapping}), selected triplets $X$ (\cref{sec:convert_triplets})
		\State \textbf{Output:} Matched and filtered triplets
		\Procedure{ApplyMatching}{$L, X$}
		\State $X' \gets \emptyset$
		\ForAll{$t$ in $X_k$}
			\If{$t_{sbj} \in L$ and $t_{obj} \in L$}
				\State $t' \gets (L[t_{sbj}], t_{predicate}, L[t_{obj}])$
				\State $X' \gets X' \cup \{t'\}$
			\EndIf
		\EndFor
		\State \textbf{return} $X'$
		\EndProcedure
	\end{algorithmic}
\end{algorithm}

The instance matching procedure consists of two parts.

First, a mapping from predicted instances to ground truth instances is generated (\cref{alg:get_mapping}). For each predicted instance, the ground truth instance with the same class label and the highest overlap is selected if the overlap surpasses a fixed threshold. Usually, the threshold is set to 0.5.
To measure overlap between instances, IoU is used on either segmentation masks or bounding boxes depending on the type of dataset. \pkg{} supports both modes.
Only a single predicted instance is allowed to be matched with a ground truth instance. Therefore, all instances that share the same matched ground truth are compared and the one with the highest overlap is selected. All other predicted instances are discarded and are matched to no ground truth at all.

Next, the mapping is used to convert a set of selected output triplets to a set of matched ground truth triplets (\cref{alg:apply_match}). During this process, some triplets may not be fully matched and are thus discarded. Hence, it is not unusual that the set of selected triplets shrinks during the matching step.

\subsection{Recall@k ($R@k$)}
\label{sec:rk}

Recall@k is the original metric for scene graph generation \cite{first_scenegraph}. It measures how good a model can retrieve ground truth triplets with the top $k$ predictions. To calculate recall, only true positives and false negatives have to be counted. Therefore, only positive ground truth annotations are required to calculate the score, which is ideal for scene graph generation, because datasets are lacking explicit negative ground truth annotation.


\subsubsection{Triplet Selection}
\label{sec:rk-triplet}

For Recall@k, a model must provide the $k$ most confident triplets, denoted as $X_k$. How these triplets are selected is up to the model. In $X_k$, no two triplets with the same subject and object are allowed.

\subsubsection{Score Definition}

To calculate the per-image score, the recall between predicted triplets $X_k$ and ground truth triplets $G$ is returned:

\begin{equation}
	\label{eq:rk}
	R@k = \frac{|G \cap X_k|}{|G|}
\end{equation}

\subsection{Mean Recall@k ($mR@k$)}
\label{sec:mrk}

\begin{algorithm}[tb]
	\caption{Mean Recall for a Single Image}
	\label{alg:meanrecall}
	\begin{algorithmic}[1]
		\State \textbf{Input:} Set of all predicate classes $P$, ground truth triplets $G$, matched triplets $X_k$ (\cref{alg:apply_match})
		\State \textbf{Output:} Mean Recall@k
		\Procedure{MeanRecall}{$P, G, X_k$}
		\State $P' \gets$ subset of $P$ that contains only predicates that exist in $G$
		\ForAll{predicate classes $p$ in $P$}
			\State $G^{(p)} \gets \{\,t \in G \mid t_{predicate} = p\,\}$
			\State $X^{(p)} \gets \{\,t \in X_k \mid t_{predicate} = p\,\}$
		\EndFor
		\State \textbf{return} $\frac{1}{|P'|} \sum\limits_{p \in P'} \frac{|G^{(p)} \cap X^{(p)}_k|}{|G^{(p)}|}$
		\EndProcedure
	\end{algorithmic}
\end{algorithm}

Mean Recall@k \cite{vctree,kern} addresses the predicate imbalance in commonly used scene graph datasets by calculating a Recall@k score for each predicate individually and averaging the scores with equal weights. This ensures that rare predicates have the same influence on the final score as common predicates. The metric is depicted in \cref{alg:meanrecall}.

\subsubsection{Triplet Selection}

Mean Recall@k uses the exact same triplet selection as Recall@k (\cref{sec:rk-triplet}).

\subsubsection{Score Definition}

Let $P'$ be the subset of $P$ that contains only predicate classes that exist in the ground truth triplets $G$. To calculate the score for each predicate individually, the set of ground truth triplets $G$ is split into individual sets $G^{(p)}$ for each predicate $p \in P'$. The same is done for the set of selected triplets $X_k$, which is split into $X_k^{(p)}$. Note that each $X_k^{(p)}$ usually contains less then $k$ elements, because the $k$ elements are distributed among the individual $X_k^{(p)}$.

The per-image score is calculated as shown in \cref{eq:mrk}.

\begin{equation}
	\label{eq:mrk}
	mR@k = \frac{1}{|P'|} \sum_{p \in P'} \frac{|G^{(p)} \cap X^{(p)}_k|}{|G^{(p)}|}
\end{equation}

\subsection{Pair Recall@k ($PR@k$)}
\label{sec:prk}

Pair Recall@k \cite{pairnet} ignores the predicted predicate from the model and just measures how good a model can detect the existence of a relation.

\subsubsection{Triplet Selection}

Pair Recall@k uses the exact same triplet selection as Recall@k (\cref{sec:rk-triplet}).

\subsubsection{Score Definition}

Pair Recall@k ignores the predicate labels. $G'$ denotes the set of subject-object pairs that is derived from $G$ by stripping the predicate label from the original triplet. The same is done to the selected triplets $X_k$, which results in $X'_k$. Then, Pair Recall@k is calculated similarly to Recall@k, as defined in \cref{eq:prk}.

\begin{equation}
	\label{eq:prk}
	PR@k = \frac{|G' \cap X'_k|}{|G'|}
\end{equation}

\subsection{No Graph Constraint Recall@k ($ngR@k$)}
\label{sec:ngrk}

No Graph Constraint Recall@k (ngR@k) \cite{pixel2graph,motifs} allows any number of triplets for the same subject-object pair in the set of selected triplets as long as their predicates are different. Therefore, $ngR@k$ can evaluate second guesses from scene graph models.

In contrast to SGG, PSGG methods have not been evaluated on this metric. We fill that gap and provide a complete evaluation over all discussed PSGG methods.


\subsubsection{Triplet Selection}
\label{sec:ngrk-triplets}

$ngR@k$ is less restrictive than $R@k$ and allows any number of triplets that share the same subject-object combination as long as their predicates are different.




\subsubsection{Score Definition}

If the triplets are selected as described above, the score can be calculated like Recall@k (\cref{sec:rk}) but with a different $X_k$.

\subsection{Mean No Graph Constraint Recall@k ($mNgR@k$)}
\label{sec:mngrk}

$mNgR@k$ is a variant of $ngR@k$ that weights each predicate class equally. It is defined analogously to $mR@k$.
%
%
The same triplet selection as in $ngR@k$ applies.
%
%
The score is calculated like Mean Recall@k if $X_k$ is replaced with the correct triplet selection process.


\subsection{Choice of $k$}
\label{sec:k}

Usually, a constant absolute number is chosen for $k$. Common values are 20, 50, and 100. This means that on every image, the same number of triplets is selected.

In addition, \pkg{} supports values for $k$ that are relative to the number of ground truth annotations. We denote relative $k$ with a multiplication symbol, \eg, R@$\times$10. With a relative $k$ of 1, we can evaluate how good a model performs if it can only output as many triplets as there are ground truth annotations.

\subsection{Instance Recall ($InstR$)}
\label{sec:inst_recall}

Scene graph model performance depends on the quality of the instance matching. Instance Recall measures how many instances are retrieved by the scene graph model. To calculate it, the mapping $L$ from \cref{sec:inst_match} can be used. Given a set of ground truth instances $M_{gt}$, Instance Recall calculates how many predicted instances $M_{out}$ can be matched to $M_{gt}$.

\begin{equation}
	InstR = \frac{|\{m \in M_{out} \mid L[m] \in M_{gt}\}|}{|M_{gt}|}
\end{equation}

\subsection{Metrics With $k = \infty$}

Apart from calculating Instance Recall, we can use the predicted instances of a scene graph model and calculate what $R@k$, $mR@k$, $ngR@k$, and $mNgR@k$ scores a hypothetical perfect scene graph model could achieve given these predicted instances. We use the notation of $k = \infty$ for this case.
We define $R@\infty$ as the best possible $R@k$ score given the matched instances. \Cref{alg:rkinf} shows pseudo code for $R@\infty$.
$mR@\infty$, $ngR@\infty$, and $mNgR@\infty$ are defined analogously.

\begin{algorithm}[tb]
	\caption{Recall@$\infty$ for a Single Image}
	\label{alg:rkinf}
	\begin{algorithmic}[1]
		\State \textbf{Input:} ground truth triplets $G$, mapping $L{:\,} M_{out} \rightarrow M_{gt}$ (\cref{alg:get_mapping})
		\State \textbf{Output:} Recall@$\infty$
		\Procedure{Recall@$\infty$}{$G, L$}
		\State $X' \gets \emptyset$
		\State $J \gets L^{-1}$ \Comment inverse mapping of $L$
		\ForAll{ground truth triplets $t$ in $G$}
			\If{$J[t_{sbj}] \neq null$ and $J[t_{obj}] \neq null$}
				\State $X' \gets X' \cup \{t\}$
			\EndIf
		\EndFor
		\State \textbf{return} $\frac{|G \cap X'|}{|G|}$
		\EndProcedure
	\end{algorithmic}
\end{algorithm}

\subsection{Predicate Rank ($PRank$)}
\label{sec:prank}

We introduce Predicate Rank ($PRank$) as a counterpart for Pair Recall. Whereas Pair Recall measures how good a scene graph model can identify subject-object pairs, $PRank$ measures how good a model is at choosing the correct predicate class, given a correct subject-object pair.

Within a relation, a scene graph model assigns confidence scores to each available predicate. Thus, the predicates can be sorted and the rank of the ground truth predicate can be retrieved for each relation. $PRank$ measures the average rank of the correct predicates. An average rank of 0 is best.

The related pseudo code is shown in \cref{alg:predicate_rank}. For $PRank$, all available matched triplets are used. First, we generate a lookup table $L$ that can map triplets to ranks. Next, the predicted ranks for all ground truth triplets are determined and averaged per predicate class. If a ground truth triplet could not be matched, the correct predicate cannot be determined. Consequently, the triplet is skipped in the calculation.

\begin{algorithm}[tb]
	\caption{Predicate Rank}
	\label{alg:predicate_rank}
	\begin{algorithmic}[1]
		\State \textbf{Input:} Set of all predicate classes $P$, ground truth triplets $G$, ordered sequence of all matched triplets $X$
		\State \textbf{Output:} Predicate Rank
		\Procedure{PredicateRank}{$P, G, X, rank$}
		\ForAll{predicates $p$ in $P$}
			\State $G^{(p)} \gets \{t \in G \mid t_{predicate} = p\}$
		\EndFor
		\State Initialize $L$ with $L[t] = null$ $\forall t \in X$
		\State Initialize $C$ with $C[t_{sbj}, t_{obj}] = 0$ $\forall t \in X$
		\ForAll{predicted triplets $t$ in $X$} \Comment Sorted
			\State $L[t] \gets C[t_{sbj}, t_{obj}]$
			\State increment $C[t_{sbj}, t_{obj}]$ by 1
		\EndFor
		\State $R \gets \emptyset$
		\ForAll{predicates $p$ in $P$}
			\State $R^{(p)} \gets \emptyset$
			\ForAll{triplets $t$ in $G^{(p)}$}
				\If{$L[t] \neq null$}
					\State $R^{(p)} \gets R^{(p)} \cup \{L[t]\}$
				\EndIf
			\EndFor
			\State $R \gets R \cup \left\{\overline{R^{(p)}}\right\}$ \Comment Add average of $R^{(p)}$
		\EndFor
		\State \textbf{return} $\overline{R}$ \Comment Return average of $R$
		\EndProcedure
	\end{algorithmic}
\end{algorithm}

\section{Implementation}

Our metrics implementation \pkg{} is designed as a standalone lightweight utility and library and is available as an easy to install pip package.
We have reduced the number of total dependencies to a minimum: \emph{NumPy} \cite{numpy} for efficient calculations, \emph{Pillow} \cite{pillow} to load ground truth segmentation masks that are stored as PNG images, \emph{tifffile} \cite{tifffile} to load TIFF files, and \emph{imagecodecs} \cite{imagecodecs} for LZMA or Deflate compression. There are no dependencies to large machine learning frameworks. Consequently, \pkg{} can be easily integrated into existing code bases and it is likely that newer dependency versions can be upgraded without issues.

Furthermore, we prioritized readability and ease of modification when developing \pkg{}. Our implementation uses less boilerplate code and contains less than 700 lines of code whereas the implementation from \cite{psg} contains more than 1500 lines even though we include additional metrics.


\subsection{File Format}
\label{sec:format}

Many existing scene graph model implementations use custom file formats to store output data. To encourage interoperability between methods, we design a file format for model outputs that is independent of the underlying software stack and is easy to create. In addition, we provide utility scripts to convert existing output file formats to our new standardized format.

\subsubsection{Triplets File}
\label{sec:format-triplets}

Triplet outputs are stored as a JSON \cite{JSON} file. JSON files can be easily inspected with a text editor and are widely supported by various tools and programming languages. In many scene graph method implementations, the Pickle \cite{pickle_v4} format is used to store outputs. This format is Python-specific and not widely adopted outside the Python ecosystem. It is not self-contained if custom data structures are serialized. This means that the reader would have to have access to the same class definitions as the writer. On top, Pickle is not secure and allows arbitrary code execution. JSON on the other hand is more secure against such attacks. Additionally, it is self-contained and can be easily read from a completely different code base.

\Cref{lst:tripletfile} shows an example triplet output file.
The \mintinline{json}{"version"} element is used to recognize possible future file format versions and should always be set to 1 at the moment.
For each processed image, an entry in the \mintinline{json}{"images"} list is added. Each entry contains an \mintinline{json}{"id"}, which refers to the image id in the ground truth annotation file.
\mintinline{json}{"seg_filename"} is a relative path to the respective TIFF file that contains the predicted segmentation masks.
\mintinline{json}{"instances"} is a list that contains information about the predicted instances. Each layer in the TIFF file corresponds to an entry in the instances list. Layer 0 corresponds to entry 0, layer 1 to entry 1, and so on. \mintinline{json}{"bbox"} is the bounding box, defined in $xyxy$ format. It is used if no segmentation masks are available during evaluation. \mintinline{json}{"category"} is the 0-based class index of the instance.
The \mintinline{json}{"triplets"} list is a list of \emph{subject-object-predicate} triplets, ordered by decreasing confidence. \emph{Subject} and \emph{object} are indices that refer to items in the instances list. \emph{Predicate} is the 0-based predicate class label that was inferred for the given triplet. Triplets with a higher confidence must come first. There is no limit to the length of the triplets list and \pkg{} automatically selects the correct subset of triplets based on the ordering.

\begin{listing}
	\caption{Example triplet output file}
	\label{lst:tripletfile}
\begin{minted}[tabsize=2, fontsize=\small]{json}
{
	"version": 1,
	"images": [
		{
			"id": 123,
			"seg_filename": "seg_file.tiff",
			"instances": [
				{
					"bbox": [1, 22, 333, 44.4],
					"category": 2
				},
				// more instances ...
			],
			"triplets": [
				[0, 3, 34],
				[2, 0, 13],
				// more triplets, ordered
				// by descending confidence ...
			]
		},
		// more images ...
	]
}
\end{minted}
\end{listing}

\subsubsection{Segmentation Masks}
\label{sec:format-masks}

For panoptic scene graph generation, additional segmentation masks are required. We use TIFF \cite{TIFFSpec} images with Deflate \cite{deflate_rfc1951} compression. Optionally, other compression methods like LZMA \cite{lzma} can be used too which are slower but more effective. TIFF files allow an arbitrary number of layers and can therefore support any number of overlapping segmentation masks in contrast to PNG files. HiLo, Pair-Net, PSGTR, and PSGFormer all output overlapping masks. TIFF files are widely supported and have a much smaller size than storing the overlapping masks as raw NumPy arrays.

\subsection{Benchmarking Service}

To further encourage a fair comparison between scene graph methods and improve visibility of new methods, we build a benchmarking service that runs \pkg{} under the hood. We believe that a common platform to compare recent advances in scene graph generation is beneficial to the community. The URL and server-side code can be found here: \url{https://lorjul.github.io/sgbench/}.

\begin{figure}[tb]
	\centering
	\fbox{\includegraphics[width=0.97\linewidth]{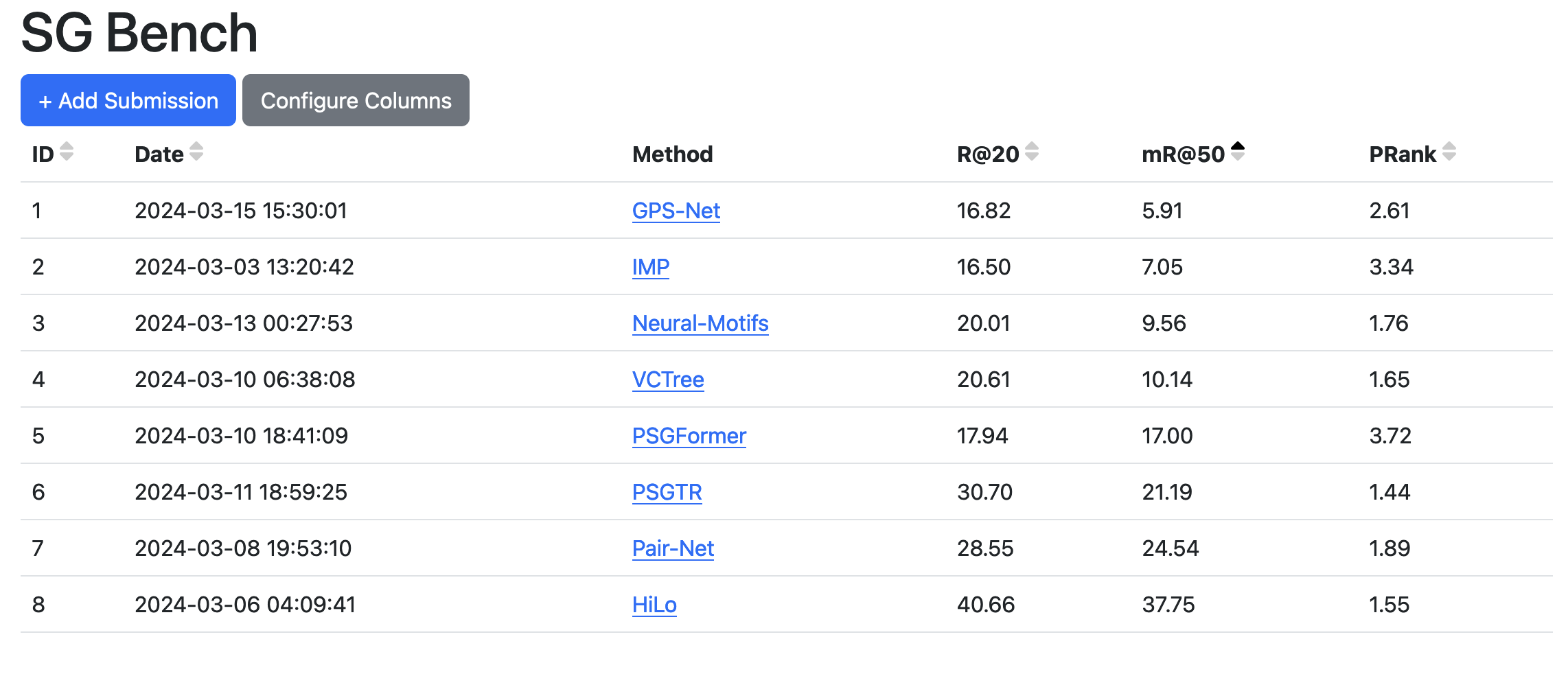}}
	\caption{Screenshot of the benchmarking service interface.}
	\label{fig:server-screenshot}
\end{figure}

To upload generated scene graphs to the benchmarking server, users have to bring their model output to the format described in \cref{sec:format}. The format we discussed consists of multiple files (one triplet file and several segmentation mask files). For submission, all these files have to be combined into a ZIP-archive. This archive can be uploaded to the benchmarking server.

The server then schedules the submission file for evaluation and adds the results to a publicly facing leaderboard. \Cref{fig:server-screenshot} shows how the leaderboard looks like. Users can add hyperlinks to their submitted results to refer to their research. We have already added the results for all discussed methods in this paper and linked to the respective sources.

To ensure a fair competition on the leaderboard, we add an additional \emph{approved} leaderboard where only we have upload access. We act as a central authority that periodically evaluates new published scene graph methods and uploads them to that leaderboard.

We believe that our benchmarking service can serve as a reference to compare scene graph methods on equal terms and as a place to increase visibility of published methods.

\section{Experiments}

\subsection{Benchmark of Existing PSGG Methods}
\label{sec:psgg_results}

In this section, we evaluate existing scene graph methods using the described metrics and compare the results. For IMP \cite{imp}, Neural Motifs \cite{motifs}, GPS-Net \cite{gpsnet}, VCTree \cite{vctree}, and HiLo \cite{hilo}, we use the published weights from the authors. For Pair-Net \cite{pairnet}, no weights are provided and we train a new model using the default instructions. We convert the output files from all scene graph methods to our described format. Finally, we evaluate using all the described metrics on the converted outputs and compare the results.

\begin{table}[tb]
	\centering
	\resizebox{\linewidth}{!}{%
	\begin{tabular}{lrrrrr}
		\toprule
		Method & R@20 $\uparrow$ & R@50 $\uparrow$ & R@100 $\uparrow$ & R@$\times$1 $\uparrow$ & R@$\times$10 $\uparrow$ \\
		\midrule
		IMP & 16.5 & 18.1 & 18.6 & 12.1 & 18.4 \\
		Neural-Motifs & 20.0 & 21.7 & 22.0 & 15.1 & 21.8 \\
		GPS-Net & 16.8 & 18.6 & 19.2 & 12.2 & 18.8 \\
		VCTree & 20.6 & 22.1 & 22.5 & 16.0 & 22.3 \\
		PSGTR & 30.7 & 35.1 & 35.6 & 19.8 & 35.2 \\
		PSGFormer & 17.9 & 19.6 & 20.0 & 12.8 & 19.7 \\
		Pair-Net & 28.5 & 34.3 & 36.8 & 18.4 & 34.9 \\
		HiLo & \textbf{40.7} & \textbf{48.7} & \textbf{51.4} & \textbf{25.2} & \textbf{49.4} \\
		\bottomrule
		\end{tabular}
	}
	\caption{Comparison of Recall values for different panoptic scene graph methods, evaluated on the PSG datast \cite{psg}. Higher values are better. A multiplication symbol ($\times$) indicates a $k$ that is relative to the number of ground truth annotations (see \cref{sec:k}).}
	\label{tab:recall}
\end{table}

\begin{figure}
	\centering
	\includegraphics[width=\linewidth]{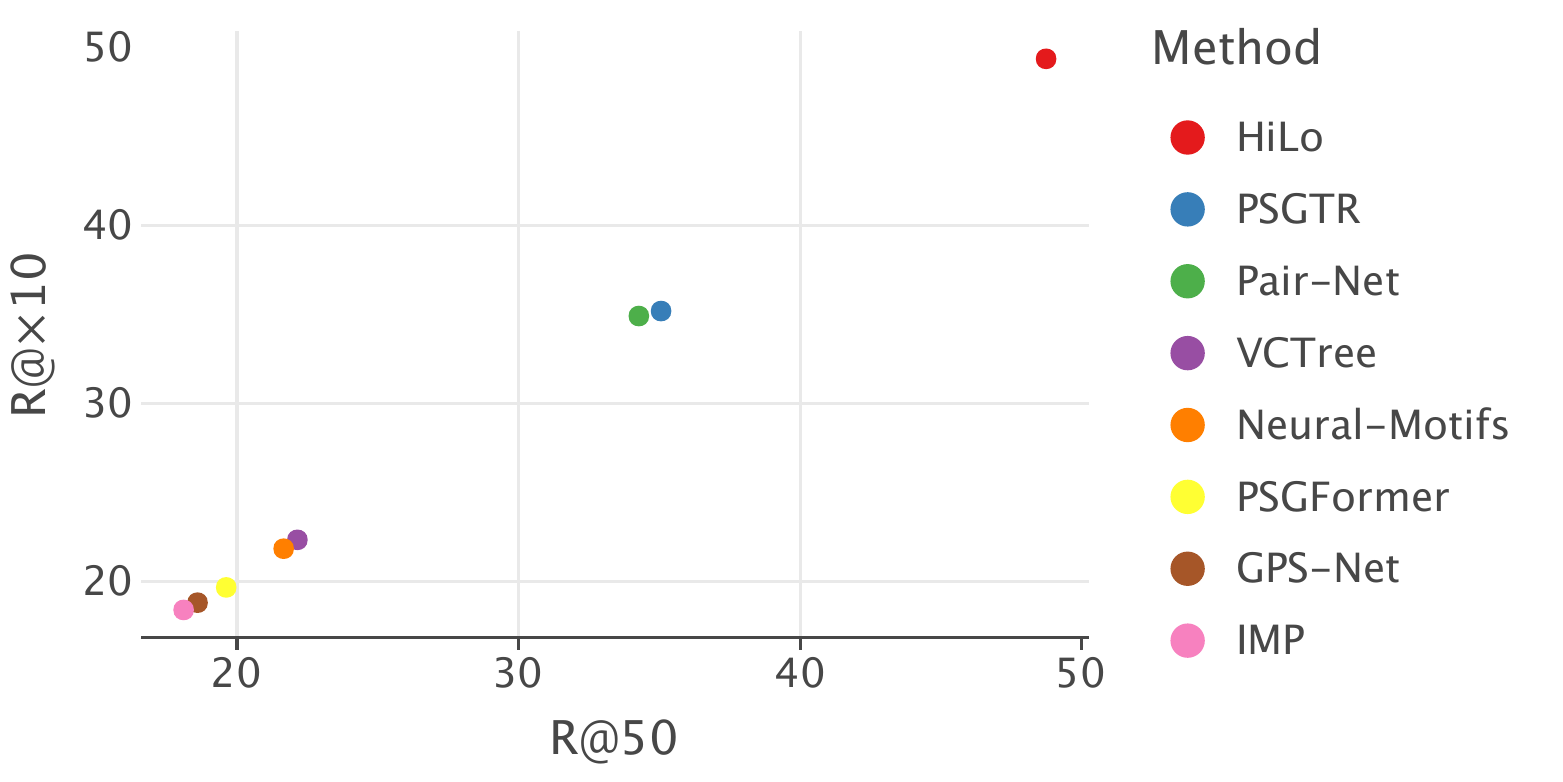}
	\caption{Absolute vs relative choice of $k$ for $R@k$ scores. $R@50$ and $R@{\times}10$ are correlated with a correlation factor of about 0.9998. This shows that both choices of $k$ are equally suited for evaluation. However, a relative $k$ is independent of the dataset.}
	\label{fig:absvsrel}
\end{figure}

\Cref{tab:recall} shows Recall@k values for various $k$. Apart from the common $k \in \{20, 50, 100\}$, we include two relative values for $k$. For $R@{\times}10$, it is allowed to select 10 triplets per ground truth annotation. For $R@{\times}1$, the number of output triplets has to be the same as the number ground truth triplets. As we show in \cref{fig:absvsrel}, similar Recall@k scores are achieved between absolute and relative $k$. However, the Recall@k scores with an absolute $k$ depend on the number of ground truth annotations in the dataset. Recall@k scores with relative $k$ are more comparable across different datasets.

\begin{table}[tb]
	\centering
	\resizebox{\linewidth}{!}{%
	\begin{tabular}{lrrrrr}
		\toprule
		Method & mR@20 $\uparrow$ & mR@50 $\uparrow$ & mR@100 $\uparrow$ & mR@$\times$1 $\uparrow$ & mR@$\times$10 $\uparrow$ \\
		\midrule
		IMP & 6.5 & 7.0 & 7.2 & 5.1 & 7.2 \\
		Neural-Motifs & 9.1 & 9.6 & 9.7 & 7.8 & 9.6 \\
		GPS-Net & 5.4 & 5.9 & 6.1 & 4.3 & 6.0 \\
		VCTree & 9.7 & 10.1 & 10.2 & 8.1 & 10.2 \\
		PSGTR & 18.1 & 21.2 & 21.4 & 10.6 & 21.3 \\
		PSGFormer & 14.8 & 17.0 & 17.6 & 9.8 & 17.1 \\
		Pair-Net & 21.5 & 24.5 & 25.6 & 15.9 & 24.9 \\
		HiLo & \textbf{29.7} & \textbf{37.8} & \textbf{41.0} & \textbf{18.8} & \textbf{39.6} \\
		\bottomrule
		\end{tabular}
	}
	\caption{Comparison of Mean Recall scores. Higher values are better. The column names are defined analogously to \cref{tab:recall}.}
	\label{tab:meanrecall}
\end{table}

\begin{table*}[tb]
	\centering
	\resizebox{0.75\linewidth}{!}{%
	\begin{tabular}{lrrrrrr}
		\toprule
		Method & mR@50 $\uparrow$ & $mNgR@50$ $\uparrow$ & PR@50 $\uparrow$ & $PRank$ $\downarrow$ & $mR@\infty$ $\uparrow$ & $InstR$ $\uparrow$ \\
		& \cref{sec:mrk} & \cref{sec:mngrk} & \cref{sec:prk} & \cref{sec:prank} & \cref{sec:k} & \cref{sec:inst_recall} \\
		\midrule
		IMP & 7.0 & 15.7 & 39.4 & 3.3 & 45.9 & 52.1 \\
		Neural-Motifs & 9.6 & 23.0 & 40.0 & 1.8 & 45.9 & 52.1 \\
		GPS-Net & 5.9 & 15.7 & 39.0 & 2.6 & 45.9 & 52.1 \\
		VCTree & 10.1 & 23.6 & 40.1 & 1.6 & 45.9 & 52.1 \\
		PSGTR & 21.2 & 22.8 & 49.2 & \textbf{1.4} & 66.2 & 65.9 \\
		PSGFormer & 17.0 & 14.5 & 27.8 & 3.7 & 54.3 & 55.1 \\
		Pair-Net & 24.5 & 21.9 & \textbf{60.9} & 1.9 & \textbf{76.6} & \textbf{75.0} \\
		HiLo & \textbf{37.8} & \textbf{37.4} & 57.6 & 1.5 & 75.6 & 73.1 \\
		\bottomrule
		\end{tabular}
	}
	\caption{Evaluation scores for various metrics on the PSG dataset \cite{psg}. For all metrics except $PRank$, higher values are better.}
	\label{tab:pairrecall}
\end{table*}

\Cref{tab:meanrecall} shows Mean Recall@k values. We use the same values for $k$ as in \cref{tab:recall}. HiLo outperforms all other methods across all metrics.

The remaining metrics are shown in \cref{tab:pairrecall}. We add $mR@50$ to the table for reference. HiLo outperforms all other methods on the main $mR@50$ and $mNgR@50$ metrics. Pair-Net is trained to improve $PR@50$ and it achieves state-of-the-art scores there as well as on $mR@\infty$ and $InstR$. However, it achieves a lower $PRank$ than HiLo which explains why Pair-Net is still inferior on the overall $mR@50$ metric. The best $PRank$ is achieved by PSGTR, but it has a comparatively low $PR@50$ which indicates why performance on $mR@50$ is not competitive.
It is worth noting that the worst $PRank$ score of 3.3 by IMP is still an acceptable value, supporting the hypothesis by \citeauthor{pairnet} \cite{pairnet} that focusing on Pair Recall should be explored more in future research.

\begin{figure}[tb]
	\centering
	\includegraphics[width=\linewidth]{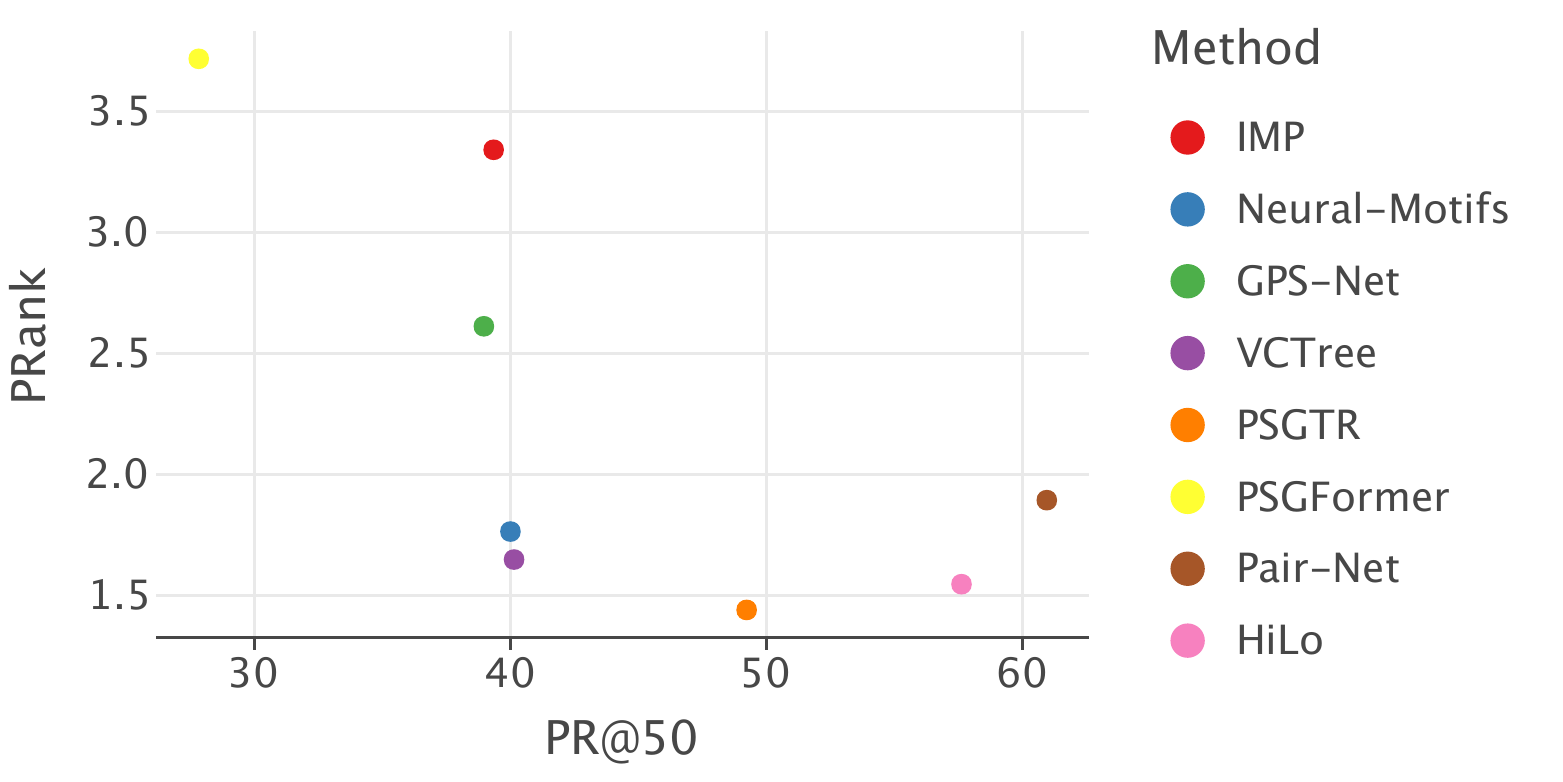}
	\caption{Pair Recall@50 ($PR@50$) compared to Predicate Rank ($PRank$). A higher $PR@50$ and a lower $PRank$ is better. A better Predicate Rank does not necessarily result in a better Pair Recall. For example PSGTR has a better $PRank$ than HiLo but a worse PR@50.}
	\label{fig:pr_prank}
\end{figure}

In \cref{fig:pr_prank}, we visualize the interaction between $PR@50$ and $PRank$. Methods that perform good on one of the metrics don't necessarily improve on the other. For example, Pair-Net has the best $PR@50$ but a $PRank$ that is worse than half of the discussed scene graph methods. Pair Recall and Predicate Rank can be used as additional metrics to gain valuable insights into model performance apart from the commonly used Mean Recall@k metrics.

\subsection{Comparison to Existing Implementation}

All discussed scene graph methods use the same implementation to calculate the metrics \cite{psg}. In this section, we will refer to that implementation as OpenPSG, based on the corresponding code repository name.

The evaluation and metrics code of OpenPSG is tightly integrated into the rest of the code base. This makes it very difficult to use as a standalone utility in a separate environment. \pkg{} on the other hand has only four dependencies in total, is designed to be integrated into existing pipelines, and can be easily installed via pip.

\pkg{} supports multiple CPU cores and can thus run the evaluation much faster. While OpenPSG takes about 66 seconds to process HiLo output, \pkg{} does it in 20 seconds. Note that \pkg{} is not only faster but also calculates additional metrics that OpenPSG does not.

OpenPSG supports a slim output format via the \texttt{--submit} flag. However, this option does not allow overlapping masks which are required for PSGTR, PSGFormer, Pair-Net, and HiLo. Therefore, the only option is to use the Pickle output format. This format takes 117 GB of disk space for Pair-Net. \pkg{} requires only 761 MB, while still containing all the necessary information for evaluation.

\section{Conclusion}

This paper has filled a significant gap in the scene graph literature by providing a thorough review and precise definitions of existing scene graph metrics.
By formalizing these metrics, we have established a solid foundation for evaluating scene graph generation models that can serve both as a reference and an introduction to scene graph generation metrics.

Furthermore, we have implemented a Python package \pkg{} that is lightweight, efficient, and easy to use.
We presented our new benchmarking service and we aim to establish it as a central place to compare scene graph generation methods for various tasks.

{
    \small
    \bibliographystyle{ieeenat_fullname}
    \bibliography{main}
}


\end{document}